\documentclass[final,5p,times,authoryear]{elsarticle}

\usepackage[utf8]{inputenc}
\usepackage[T1]{fontenc}
\usepackage{amsmath,amssymb}
\usepackage{booktabs}
\usepackage{multirow}
\usepackage{array}
\usepackage{graphicx}
\usepackage{caption}
\usepackage{float}
\usepackage{placeins}
\usepackage{dblfloatfix}
\usepackage{flafter}

\usepackage{enumitem}
\usepackage{xurl}
\usepackage{hyperref}
\hypersetup{
  hidelinks,
  pdfauthor={Lin Jiang, Qingshan She, Jiale Xu, Haiqi Xu, Duanpo Wu, Zhenzhong Kuang},
  pdftitle={Subject-Aware Multi-Granularity Alignment for Zero-Shot EEG-to-Image Retrieval},
  pdfsubject={},
  pdfkeywords={}
}
\usepackage{microtype}
\journal{Expert Systems with Applications}

\biboptions{authoryear,sort}

\newcommand{\softmax}{\operatorname{softmax}}
\newcommand{\Bern}{\operatorname{Bernoulli}}

\begin{document}

\begin{frontmatter}

\title{Subject-Aware Multi-Granularity Alignment for Zero-Shot EEG-to-Image Retrieval\tnoteref{funding}}

\author[aff1]{Lin Jiang}
\ead{252060400@hdu.edu.cn}
\author[aff1]{Qingshan She\corref{cor1}}
\ead{qsshe@hdu.edu.cn}
\author[aff1]{Jiale Xu}
\ead{2286676614@qq.com}
\author[aff1]{Haiqi Xu}
\ead{232060231@hdu.edu.cn}
\author[aff1]{Duanpo Wu}
\ead{41711@hdu.edu.cn}
\author[aff2]{Zhenzhong Kuang}
\ead{zzkuang@hdu.edu.cn}

\cortext[cor1]{Corresponding author: Qingshan She.}
\tnotetext[funding]{This work was supported in part by Zhejiang Provincial Natural Science Foundation of China (No. LZ26F010007) and National Natural Science Foundation of China (No. 62371172).}

\affiliation[aff1]{organization={School of Automation, Hangzhou Dianzi University; Zhejiang Provincial Key Laboratory of Brain Computer Collaborative Intelligence Technology and Applications},
            city={Hangzhou},
            postcode={310018},
            state={Zhejiang},
            country={China}}

\affiliation[aff2]{organization={College of Computer Science and Technology, Hangzhou Dianzi University},
            city={Hangzhou},
            postcode={310018},
            state={Zhejiang},
            country={China}}

\begin{abstract}
Decoding visual content from electroencephalography (EEG) is important for understanding neural visual representations and developing non-invasive brain--computer interfaces. Existing approaches mainly improve EEG representation learning and cross-modal alignment while treating pretrained visual representations as fixed supervision targets. However, pretrained vision models organize information hierarchically, with different depths encoding complementary structural and semantic information, and the visual granularity most compatible with EEG may vary across subjects. To address this issue, we propose Subject-Aware Multi-Granularity Alignment (SAMGA), which makes visual-target construction an explicit part of EEG--visual alignment. SAMGA constructs adaptive visual supervision from multiple intermediate representations and models EEG-compatible visual granularity through a global granularity prior with subject-dependent residual calibration, enabling subject-aware training and subject-agnostic inference. Based on the resulting adaptive target, a coarse-to-fine alignment strategy first organizes global cross-modal geometry and then refines instance-level retrieval discrimination. On THINGS-EEG, SAMGA improves Top-1 retrieval accuracy over the strongest competing method by 8.7 percentage points under intra-subject evaluation and 12.0 percentage points under leave-one-subject-out evaluation. These results support a broader view of neural--visual alignment, in which performance depends not only on how neural representations are mapped, but also on what visual representations define their supervision.
\end{abstract}


\end{frontmatter}

\section{Introduction}

Visual decoding from human brain activity provides a computational framework for characterizing neural representations and inferring perceptual content from measured brain responses \citep{r4,r23,r20,r5}. The availability of large-scale natural-image EEG datasets and pretrained visual and vision--language models has extended this paradigm to semantically structured visual spaces \citep{r12,r10,r40,r6}. In particular, NICE established contrastive EEG--image alignment on THINGS-EEG, demonstrated zero-shot retrieval of object concepts excluded from model training, and provided a representative paradigm for EEG-to-image retrieval at scale \citep{r13}. Subsequent work has enriched this paradigm through improved neural representation learning, multimodal semantic supervision, latent-space regularization, and cross-modal alignment objectives \citep{r16,r14,r15,r18}.

Much of this progress has centered on EEG representation learning and cross-modal alignment within pretrained visual spaces. By contrast, the visual target is commonly specified in advance by the chosen pretrained representation. This design choice directly determines which visual information the EEG encoder is encouraged to recover. Deep visual models organize information hierarchically, progressively transforming local visual structure into higher-level and increasingly abstract representations \citep{r37,r21}. Multi-level deep features also exhibit distinct correspondence with neural visual representations and have been exploited for brain-based visual decoding \citep{r20,r25}. Visual supervision is therefore not merely an endpoint of alignment; its representational granularity is itself part of the alignment problem.

\begin{figure*}[!t]
\centering
\includegraphics[width=0.96\textwidth]{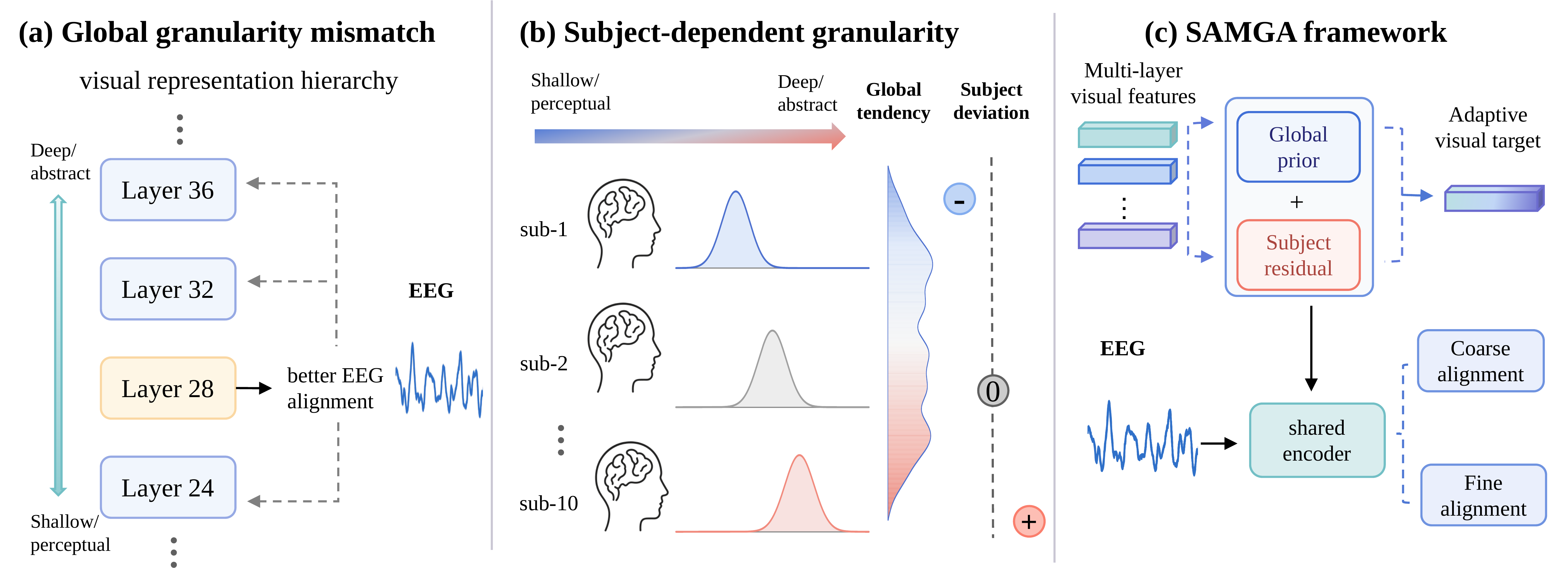}
\caption{Motivation and overview of SAMGA. (a) Global granularity mismatch motivates alignment with intermediate rather than overly abstract visual representations. (b) EEG-compatible visual granularity varies across subjects around a shared granularity tendency. (c) SAMGA constructs adaptive visual supervision through a global prior and subject residual calibration, followed by coarse-to-fine EEG--image alignment.}
\label{fig:motivation}
\end{figure*}

Recent neural decoding studies provide direct support for this perspective. UBP modifies image-side information through uncertainty-aware blurring of high-frequency visual details, reducing the impact of mismatch between visual stimuli and brain signals during alignment \citep{r17}. More fundamentally, Shallow Alignment identifies a granularity mismatch between neural representations and highly abstract final-layer features of deep visual models, demonstrating that intermediate representations can provide substantially more effective supervision for neural decoding \citep{r22}. Together with established evidence for hierarchical correspondence between computational and biological visual representations \citep{r21,r20}, these findings elevate visual depth from a backbone configuration to an explicit supervision-design variable. Collectively, these results broaden neural--visual alignment beyond neural mapping to include the design of the visual representation that defines its supervision.

Granularity-aware supervision raises a further issue when EEG observations from different individuals are considered. Existing granularity-aware formulations still employ visual granularity as a globally shared choice across subjects \citep{r22}, whereas neural observations retain substantial inter-individual variability despite common representational organization across individuals \citep{r23,r10,r5}. Related multi-subject visual decoding studies similarly combine shared representations with subject-specific mappings or adaptation mechanisms \citep{r32,r33,r34,r35}. These observations motivate a subject-dependent view of granularity mismatch, in which the visual representation most compatible with EEG varies across subjects around a global granularity tendency.

We formulate this as subject-dependent granularity mismatch, in which EEG-compatible visual granularity comprises a global granularity tendency and subject-dependent deviations. This formulation treats subject-dependent information as a calibration signal for visual supervision, while the global granularity tendency preserves transferable structure across subjects. As illustrated in Fig.~\ref{fig:motivation}, this perspective establishes a progression from fixed visual targets, to globally optimized granularity, and finally to subject-aware target construction around a global granularity prior. This progression yields the central design principle of our framework: subject-aware training with subject-agnostic inference.

Building on this formulation, we propose Subject-Aware Multi-Granularity Alignment (SAMGA) for zero-shot EEG-to-image retrieval. SAMGA learns visual supervision from multiple intermediate representations rather than selecting a single visual target in advance. It models EEG-compatible visual granularity as a shared tendency with subject-dependent residual deviations, allowing differences in visual-depth compatibility across subjects to calibrate target construction during training while retaining the learned global tendency at inference. Beyond target construction, SAMGA further organizes the resulting shared EEG--image space through a coarse-to-fine cross-modal strategy. Coarse alignment first establishes global cross-modal structure, followed by fine alignment that strengthens instance-level discrimination within the learned shared space. In this way, SAMGA learns how visual supervision should be constructed rather than merely learning an EEG mapping toward a predefined visual target, while progressively organizing the resulting cross-modal representation for discriminative zero-shot retrieval.

The main contributions of this work are summarized as follows:

\begin{enumerate}[label=\arabic*)]
\item We formulate subject-dependent granularity mismatch as an explicit dimension of neural--visual alignment, extending granularity-aware decoding from globally shared visual-depth selection to subject-dependent visual-target modeling.
\item We develop a subject-aware multi-granularity target construction framework that learns visual supervision rather than treating it as predefined. A global granularity prior captures shared granularity structure, while subject residuals calibrate supervision during training; only the global prior is retained at inference, realizing subject-aware training with subject-agnostic inference.
\item We introduce a coarse-to-fine cross-modal alignment strategy that progressively organizes the shared EEG--image space from global cross-modal structure to instance-level discrimination, supporting discriminative zero-shot retrieval.
\item Experiments across both intra-subject and leave-one-subject-out settings demonstrate consistent improvements in zero-shot EEG-to-image retrieval. Ablation and visual-depth analyses further provide empirical evidence for the learned global granularity tendency and subject-dependent granularity behavior.
\end{enumerate}

\section{Related Work}

\subsection{EEG-based Visual Decoding}

EEG-based visual decoding has progressively evolved from early object recognition with learned neural representations toward retrieval and reconstruction in semantically structured visual spaces \citep{r41,r42}. Multimodal brain-decoding studies have shown that visual and linguistic representations can provide complementary semantic supervision beyond conventional category-level prediction \citep{r6}. The availability of large-scale natural-image EEG data and pretrained visual and vision--language models has further enabled neural responses to be associated with richer visual representations and previously unseen object concepts. An early step toward this setting formulated EEG-based image retrieval through deep metric learning in a shared neural--visual embedding space \citep{r43}. Building on large-scale natural-image EEG data and pretrained visual encoders, NICE introduced contrastive EEG--image alignment for zero-shot object retrieval on THINGS-EEG, establishing a representative framework for subsequent EEG--image retrieval studies \citep{r13}.

Subsequent studies have extended this paradigm through richer semantic supervision, generative priors, and more structured latent-space alignment. NICE++ incorporates language-guided contrastive supervision to strengthen semantic correspondence between EEG and visual representations \citep{r14}, while CognitionCapturer exploits complementary multimodal information beyond paired EEG and images and Neural-MCRL further strengthens neural--visual correspondence through multimodal semantic bridging and cross-attention \citep{r44,r45}. E2IVAE emphasizes cross-modal latent-distribution alignment together with an interpretable algorithm-unrolled EEG encoder \citep{r15}. NeuroBridge further introduces cognitive-prior augmentation and bidirectional semantic alignment to strengthen EEG--image correspondence \citep{r18}. More recently, multimodal visual decoding has also incorporated EEG, image, and text representations with dynamic modality balancing to improve the robustness of cross-modal learning \citep{r48}.

A parallel line of work couples neural representation learning with visual retrieval and reconstruction. ATM aligns EEG with CLIP representations and further exploits generative priors for visual decoding \citep{r16}. BrainVision extends EEG representation learning toward both visual-content retrieval and reconstruction \citep{r46}, whereas NeuroVision combines progressive neural encoding with cross-modal distillation to preserve semantic information during EEG-to-image reconstruction \citep{r47}. Temporally resolved visual retrieval and reconstruction have likewise been investigated from MEG using pretrained visual representations \citep{r49}. Beyond direct visual decoding, multimodal EEG representation learning has also explored adaptive fusion across EEG, audio, and visual information \citep{r50}, reflecting the broader use of complementary modalities for neural representation learning. Collectively, these studies establish pretrained semantic spaces and cross-modal representation alignment as a central paradigm for zero-shot EEG-based visual decoding.

\subsection{Cross-modal Representation Alignment and Visual Target Construction}

The use of pretrained visual spaces naturally raises the issue of which visual representation should serve as the neural supervision target. Deep visual networks encode information hierarchically, and different representation depths exhibit distinct relationships with biological visual processing. G\"u\c{c}l\"u and van Gerven \citeyearpar{r21} identified a gradient of representational complexity between deep-network layers and the human ventral visual stream, while Horikawa and Kamitani \citeyearpar{r20} demonstrated that visual features spanning multiple hierarchical levels can be decoded from brain activity. Shen et al. \citeyearpar{r25} further incorporated multi-level deep visual features as complementary constraints for brain-based image reconstruction. Together, these studies establish the representational basis for treating different visual depths as candidate supervision targets rather than assuming that a single final-layer representation is universally optimal.

Recent neural decoding methods explicitly optimize the visual side of alignment. UBP estimates uncertainty in paired visual--brain data and uses uncertainty-aware blurring to reduce the contribution of mismatched high-frequency visual information during alignment \citep{r17}. NeuroCLIP adapts CLIP-side representations through neural-aware visual prompt tuning and instance-conditioned visual modulation for EEG--image contrastive learning \citep{r19}. Despite their different mechanisms, these approaches demonstrate that the visual side of neural decoding can be actively adapted rather than treated as fixed supervision.

Shallow Alignment is most directly related to the granularity perspective considered in this work \citep{r22}. It identifies a granularity mismatch between neural signals and highly abstract final-layer visual representations and replaces conventional final-layer supervision with intermediate features that retain a different balance of structural and semantic information. Its experiments demonstrate substantial improvements across multiple visual backbones, establishing representation depth as an explicit factor in neural--visual alignment. However, granularity matching is implemented through a globally selected intermediate representation rather than subject-aware construction of a multi-layer visual target.

Using hierarchical representations is therefore not equivalent to adaptively constructing visual supervision. Existing target-side methods establish that visual depth and visual adaptation matter, but the resulting supervision remains globally defined across subjects. Subject-aware construction of visual targets from hierarchical representations has therefore received limited attention.

\subsection{Subject Variability in Neural Visual Decoding}

Subject variability is a persistent characteristic of neural visual decoding and affects the compatibility of learned representations across individuals. At the EEG level, CrossPT-EEG explicitly benchmarks cross-participant and cross-time generalization, highlighting subject and temporal variability as important evaluation dimensions for visual decoding \citep{r9}. ECHO-Net combines dynamic feature modeling and graph-based spatial representation learning to improve cross-subject EEG--vision decoding \citep{r30}. The difficulty of learning subject-independent EEG representations is also observed beyond visual retrieval: functional-connectivity-guided visual-imagery decoding has been evaluated under leave-one-subject-out generalization \citep{r51}, while domain-generalization approaches such as MI-DGHCL explicitly address cross-subject distribution shift in motor-imagery EEG \citep{r52}. These studies reinforce the importance of accounting for individual variability when learning shared EEG--visual representations.

Related fMRI research provides complementary evidence that shared neural structure and individual-specific variation can be modeled jointly. Haxby et al. \citeyearpar{r23} showed that individual visual response spaces can be transformed into a common high-dimensional representational space. More recent visual decoding studies broadly follow two directions. One direction learns shared or multi-subject representations, as exemplified by MindBridge and MindEye2, which exploit information across participants before decoding or adapting to a new individual \citep{r32,r33}. A complementary direction emphasizes subject-specific correction or explicit functional alignment, as in MindTuner and MindAligner, to accommodate individual differences within a shared decoding framework \citep{r34,r35}.

Despite their methodological differences, these approaches primarily address subject variability by modifying the neural representation, neural mapping, or adaptation process. How subject information can instead inform visual supervision construction, particularly under subject-agnostic inference, remains an open question.

\section{Methodology}
\label{sec:method}

\subsection{Problem Formulation and Framework Overview}
\label{sec:overview}

Given a training set of paired EEG responses and visual stimuli,
\[
\mathcal{D}=\left\{\left(\mathbf{x}_{n}^{E},\mathbf{x}_{n}^{I},s_n,y_n\right)\right\}_{n=1}^{N},
\]
where $\mathbf{x}_{n}^{E}\in\mathbb{R}^{C\times T}$ denotes the EEG response of subject $s_n$ to visual stimulus $\mathbf{x}_{n}^{I}$, and $y_n\in\mathcal{C}_{\mathrm{tr}}$ denotes the semantic concept associated with the $n$-th pair and is used to define the zero-shot concept split. Zero-shot EEG-to-image retrieval aims to identify the corresponding image from a candidate gallery for EEG responses associated with concepts unseen during training. Let $\mathcal{C}_{\mathrm{tr}}$ and $\mathcal{C}_{\mathrm{te}}$ denote the training and evaluation concept sets, respectively, with
\begin{equation}
\mathcal{C}_{\mathrm{tr}}\cap\mathcal{C}_{\mathrm{te}}=\varnothing.
\label{eq:zeroshot}
\end{equation}

\begin{figure*}[!t]
\centering
\includegraphics[width=0.96\textwidth]{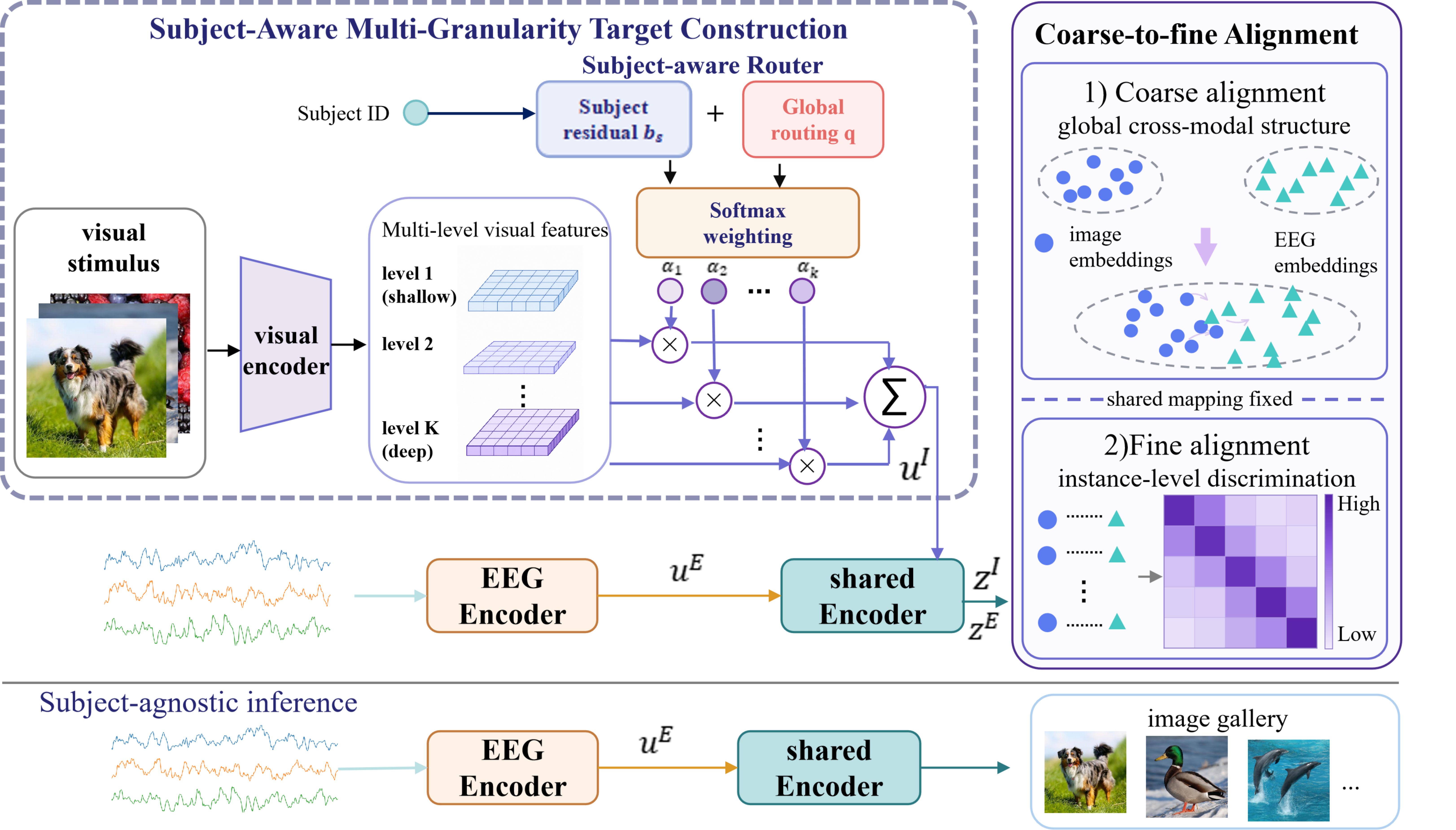}
\caption{Overall framework of SAMGA. Multi-level visual representations are adaptively combined through global--residual granularity routing during training, and the resulting visual target is aligned with EEG through a shared encoder and coarse-to-fine optimization. Inference uses the learned global granularity prior for subject-agnostic retrieval.}
\label{fig:framework}
\end{figure*}

SAMGA reframes EEG--visual alignment as a joint problem of learning what visual representation should supervise EEG representation learning and how the resulting EEG--visual correspondence should be progressively organized. Specifically, SAMGA learns an EEG-compatible visual target from hierarchical visual representations and organizes its correspondence with EEG in a shared latent space. This formulation moves beyond learning an EEG mapping to a predetermined visual target by making visual-target construction an explicit part of the alignment process.

As illustrated in Fig.~\ref{fig:framework}, SAMGA consists of two tightly coupled components: subject-aware multi-granularity target construction and shared-space coarse-to-fine alignment. The former learns adaptive visual supervision from hierarchical representations, whereas the latter organizes the resulting EEG--visual correspondence from distribution-level alignment to instance-level discrimination. Subject-dependent calibration is used only during training, while inference relies exclusively on the learned global granularity prior.

\subsection{Subject-Aware Multi-Granularity Target Construction}
\label{sec:target}

A fixed visual representation imposes a single supervision granularity on all EEG observations. We model subject-dependent visual granularity as structured deviations from a common granularity preference. Based on this formulation, SAMGA constructs visual supervision from multiple depths of a pretrained visual hierarchy and introduces a global--residual granularity routing mechanism that factorizes granularity preference into a global component and subject-dependent residuals.

Let $\mathcal{L}=\{\ell_1,\ell_2,\ldots,\ell_K\}$ denote the selected layers of a frozen visual encoder $F_{\phi}$. For image $\mathbf{x}_{n}^{I}$, the representation extracted from layer $\ell_k$ is
\begin{equation}
\mathbf{h}_{n,k}^{I}=F_{\phi}^{(\ell_k)}\!\left(\mathbf{x}_{n}^{I}\right),
\qquad k=1,\ldots,K.
\label{eq:visual_features}
\end{equation}
Different visual depths encode complementary levels of visual abstraction. Each layer representation is therefore projected into a common dimensional space,
\begin{equation}
\widetilde{\mathbf{h}}_{n,k}^{I}=p_k\!\left(\mathbf{h}_{n,k}^{I}\right),
\label{eq:projector}
\end{equation}
where $p_k(\cdot)$ denotes the lightweight projection associated with the $k$-th representation. The projected features $\{\widetilde{\mathbf{h}}_{n,k}^{I}\}_{k=1}^{K}$ constitute the hierarchical representation pool from which the visual target is constructed.

\paragraph{Global--residual granularity routing}
At the core of target construction is a granularity router that assigns adaptive contributions to the candidate visual depths. Let $\mathbf{q}\in\mathbb{R}^{K}$ denote learnable global routing logits that parameterize the global granularity prior, and let $\mathbf{b}_{s}\in\mathbb{R}^{K}$ denote the subject residual associated with training subject $s$. The global--residual factorization expresses subject-specific routing as a structured deviation from the global routing reference.

To explicitly couple subject-aware optimization with subject-agnostic deployment, SAMGA stochastically suppresses the subject residual branch through subject dropout. For sample $n$, let $r_n\sim\Bern(1-p_s)$. The routing weights are defined as
\begin{equation}
\boldsymbol{\alpha}_{n}
=
\softmax\!\left(
\frac{\mathbf{q}+r_n\mathbf{b}_{s_n}}{\tau_r}
\right),
\label{eq:router}
\end{equation}
where $\boldsymbol{\alpha}_{n}=[\alpha_{n,1},\ldots,\alpha_{n,K}]$ and $\tau_r$ denotes the routing temperature. Depending on $r_n$, the router alternates between subject-conditioned and global-only configurations during training. Consequently, the global-only routing pathway is explicitly optimized during training, matching the subject-independent routing condition used at inference.

Layer dropout is further applied to regularize visual-target composition. Let $m_{n,k}\sim\Bern(1-p_{\ell})$. To maintain a valid routing distribution, at least one candidate visual depth is retained for each sample; if all depths are dropped, the full candidate set is restored. The post-dropout routing weights are
\begin{equation}
\widehat{\alpha}_{n,k}
=
\frac{m_{n,k}\alpha_{n,k}}
{\displaystyle\sum_{j=1}^{K}m_{n,j}\alpha_{n,j}},
\label{eq:layer_dropout}
\end{equation}
Subject dropout controls the use of subject residuals, whereas layer dropout regularizes the composition of hierarchical visual supervision.

The adaptive visual target is then constructed as
\begin{equation}
\mathbf{u}_{n}^{I}
=
\sum_{k=1}^{K}
\widehat{\alpha}_{n,k}\,\widetilde{\mathbf{h}}_{n,k}^{I}.
\label{eq:visual_target}
\end{equation}
The resulting visual target is not tied to a predetermined representation depth, but is adaptively constructed from the visual hierarchy through global--residual routing.

\paragraph{Subject-agnostic target construction}
At inference, subject residuals are omitted and both dropout mechanisms are disabled. Gallery targets are constructed exclusively using the learned global granularity prior,
\begin{equation}
\boldsymbol{\alpha}^{g}
=
\softmax\!\left(\frac{\mathbf{q}}{\tau_r}\right).
\label{eq:global_prior}
\end{equation}
This requires neither subject identity nor subject-specific parameters, allowing subject information to serve purely as a training-time calibration signal. The resulting $\mathbf{u}_{n}^{I}$ subsequently serves as the visual-side target for shared-space EEG--visual alignment.

\subsection{Shared-Space Coarse-to-Fine Alignment}
\label{sec:alignment}

With the adaptive visual target established, SAMGA organizes EEG and visual representations in a shared latent space through coarse-to-fine alignment. Coarse alignment operates at the distribution level, whereas fine alignment focuses on instance-level discriminative refinement.

For EEG trial $\mathbf{x}_{n}^{E}$, the EEG representation is obtained through the EEG encoder $f_{\theta}$ and projector $p_E$,
\begin{equation}
\mathbf{u}_{n}^{E}
=
p_E\!\left(f_{\theta}(\mathbf{x}_{n}^{E})\right).
\label{eq:eeg_feature}
\end{equation}
The EEG representation $\mathbf{u}_{n}^{E}$ and adaptive visual target $\mathbf{u}_{n}^{I}$ are subsequently processed by the shared encoder $g_{\psi}$,
\begin{equation}
\mathbf{z}_{n}^{E}=g_{\psi}(\mathbf{u}_{n}^{E}),
\qquad
\mathbf{z}_{n}^{I}=g_{\psi}(\mathbf{u}_{n}^{I}).
\label{eq:shared_encoder}
\end{equation}
The shared encoder serves as a cross-modal alignment anchor, providing a common transformation under which EEG--visual correspondence is subsequently optimized.

\paragraph{Coarse alignment: distribution-level cross-modal organization}
The coarse stage establishes a shared cross-modal geometry by jointly incorporating distribution-level alignment and pairwise retrieval supervision. For a mini-batch of $M$ matched EEG--visual pairs, we define the scaled pairwise similarity as
\[
s_{ij}=\frac{\operatorname{sim}(\mathbf{z}_{i}^{E},\mathbf{z}_{j}^{I})}{\tau},
\]
where $\operatorname{sim}(\cdot,\cdot)$ denotes cosine similarity and $\tau$ is the contrastive temperature. We employ the symmetric contrastive retrieval objective
\begin{equation}
\begin{aligned}
\mathcal{L}_{E\rightarrow I}
&=-\frac{1}{M}\sum_{i=1}^{M}
\log\frac{\exp(s_{ii})}{\sum_{j=1}^{M}\exp(s_{ij})},\\
\mathcal{L}_{I\rightarrow E}
&=-\frac{1}{M}\sum_{i=1}^{M}
\log\frac{\exp(s_{ii})}{\sum_{j=1}^{M}\exp(s_{ji})},\\
\mathcal{L}_{\mathrm{ret}}
&=\frac{1}{2}\left(\mathcal{L}_{E\rightarrow I}+\mathcal{L}_{I\rightarrow E}\right).
\end{aligned}
\label{eq:retrieval_loss}
\end{equation}
The symmetric objective promotes paired EEG--visual correspondence in both retrieval directions.

To reduce modality-level distribution discrepancy, we further employ multi-kernel maximum mean discrepancy (MK-MMD) \citep{r38,r39},
\begin{equation}
\begin{aligned}
\mathcal{L}_{\mathrm{MMD}}
={}&\frac{1}{M(M-1)}\sum_{i\neq j}k(\mathbf{z}_{i}^{E},\mathbf{z}_{j}^{E})\\
&+\frac{1}{M(M-1)}\sum_{i\neq j}k(\mathbf{z}_{i}^{I},\mathbf{z}_{j}^{I})\\
&-\frac{2}{M^{2}}\sum_{i=1}^{M}\sum_{j=1}^{M}k(\mathbf{z}_{i}^{E},\mathbf{z}_{j}^{I}).
\end{aligned}
\label{eq:mmd}
\end{equation}
Here, $k(\mathbf{a},\mathbf{b})=\frac{1}{R}\sum_{r=1}^{R}k_r(\mathbf{a},\mathbf{b})$, where $k_r$ denotes an RBF kernel with bandwidth $\sigma_r$. Together, the two objectives organize the shared EEG--visual geometry by coupling distribution-level alignment with pairwise correspondence.

The coarse-stage objective at epoch $t$ is
\begin{equation}
\mathcal{L}_{\mathrm{coarse}}^{(t)}
=
\lambda_t\mathcal{L}_{\mathrm{MMD}}
+\left(1-\lambda_t\right)\mathcal{L}_{\mathrm{ret}},
\label{eq:coarse}
\end{equation}
where $\lambda_t$ decreases linearly over coarse training. Early optimization therefore emphasizes distribution-level cross-modal organization, while the contribution of paired retrieval discrimination gradually increases.

\paragraph{Fine alignment: instance-level discriminative refinement}
Once the coarse stage has established the shared EEG--visual geometry, optimization shifts toward retrieval-oriented instance refinement. The shared encoder $g_{\psi}$ is frozen, thereby fixing the alignment anchor established during coarse training, while the remaining trainable EEG-side and target-construction components continue to be optimized. MK-MMD is removed, and optimization proceeds with retrieval supervision alone under a reduced learning rate; the fine-stage objective therefore reduces to $\mathcal{L}_{\mathrm{fine}}=\mathcal{L}_{\mathrm{ret}}$.

At inference, gallery targets are constructed using the global granularity prior. Zero-shot retrieval is performed by ranking the resulting gallery embeddings according to their cosine similarity with the EEG query embedding. The complete training and inference flows are summarized in Fig.~\ref{fig:framework}.

\section{Experiments and Results}

\subsection{Datasets and Evaluation Protocols}

We evaluate SAMGA on THINGS-EEG \citep{r10}, a large-scale dataset of visually evoked EEG responses collected from ten subjects under a rapid serial visual presentation paradigm. The training split contains 1,654 object concepts, with ten images per concept and four repetitions per image. The test split contains 200 disjoint concepts, each represented by one image repeated 80 times. Following the established zero-shot retrieval protocol, all test concepts are excluded from training.

EEG signals are band-pass filtered between 0.1 and 100 Hz and segmented into epochs from 0 to 1000 ms relative to stimulus onset. Baseline correction is performed using the 200 ms pre-stimulus interval, followed by downsampling to 250 Hz and multivariate noise normalization. Repeated responses to the same test image are averaged before retrieval evaluation, following the preprocessing protocol of NICE \citep{r13}.

We further evaluate SAMGA on THINGS-MEG \citep{r36}, which contains MEG responses from four subjects recorded with 271 sensors. The original main set covers all 1,854 THINGS object concepts, with 12 distinct images presented once per concept, while a separate repeated test set contains 200 images, one per concept, each presented 12 times. Following the zero-shot retrieval protocol, the 200 concepts represented in the repeated test set are excluded from the main set, leaving 1,654 concepts for training and 200 unseen concepts for evaluation. MEG signals are segmented over 0--1000 ms, band-pass filtered between 0.1 and 100 Hz, baseline-corrected using the 200 ms pre-stimulus interval, and downsampled to 200 Hz. Channel-wise z-score normalization is performed using the mean and standard deviation computed from the training split, and the same statistics are applied to the test data. Repeated responses to each test image are averaged during evaluation.

For both datasets, retrieval is evaluated over a 200-image gallery using Top-1 and Top-5 accuracy. In the intra-subject setting, models are trained and evaluated separately for each subject. In the inter-subject setting, we adopt leave-one-subject-out evaluation, where one subject is held out for testing and the remaining subjects are used for training.

\setcounter{table}{1}
\begin{table*}[!b]
\centering
\caption{Comparison with representative methods on THINGS-EEG.}
\label{tab:main_results}
\scriptsize
\setlength{\tabcolsep}{3.2pt}
\resizebox{\textwidth}{!}{%
\begin{tabular}{llrrrrrrrrrrr}
\toprule
\multicolumn{13}{c}{\textbf{Intra-subject: train and test on one subject}}\\
\midrule
Method & Metric & Sub1 & Sub2 & Sub3 & Sub4 & Sub5 & Sub6 & Sub7 & Sub8 & Sub9 & Sub10 & Avg. \\
\midrule
NICE & Top-1 & 13.2&13.5&14.5&20.6&10.1&16.5&17.0&22.9&15.4&17.4&16.1\\
 & Top-5 & 39.5&40.3&42.7&52.7&31.5&44.0&42.1&56.1&41.6&45.8&43.6\\
ATM & Top-1 & 25.6&22.0&25.0&31.4&12.9&21.3&30.5&38.8&34.4&29.1&27.1\\
 & Top-5 & 60.4&54.5&62.4&60.9&43.0&51.1&61.5&72.0&51.5&63.5&58.1\\
UBP & Top-1 & 41.2&51.2&51.2&51.1&42.2&57.5&49.0&58.6&45.1&61.5&50.9\\
 & Top-5 & 70.5&80.9&82.0&76.9&72.8&83.5&79.9&85.8&76.2&88.2&79.7\\
NeuroBridge & Top-1 & 50.0&63.2&61.6&61.4&54.8&69.7&62.7&71.2&64.0&73.6&63.2\\
 & Top-5 & 77.6&90.6&91.1&90.0&85.0&92.9&88.8&95.1&91.0&97.1&89.9\\
Shallow Alignment & Top-1 & 75.4&87.3&82.9&79.1&74.9&90.2&79.0&86.9&81.3&89.3&82.6\\
 & Top-5 & 94.3&99.0&\textbf{98.3}&96.5&96.4&99.2&97.3&99.4&97.8&99.2&97.7\\
SAMGA & Top-1 & \textbf{85.2}&\textbf{94.4}&\textbf{91.8}&\textbf{89.7}&\textbf{86.3}&\textbf{97.2}&\textbf{89.2}&\textbf{94.8}&\textbf{88.7}&\textbf{96.4}&\textbf{91.3}\\
 & Top-5 & \textbf{95.5}&\textbf{99.8}&98.2&\textbf{98.7}&\textbf{98.4}&\textbf{99.9}&\textbf{98.7}&\textbf{99.8}&\textbf{99.2}&\textbf{99.8}&\textbf{98.8}\\
\midrule
\multicolumn{13}{c}{\textbf{Inter-subject: leave one subject out for test}}\\
\midrule
Method & Metric & Sub1 & Sub2 & Sub3 & Sub4 & Sub5 & Sub6 & Sub7 & Sub8 & Sub9 & Sub10 & Avg. \\
\midrule
NICE & Top-1 & 7.6&5.9&6.0&6.3&4.4&5.6&5.6&6.3&5.7&8.4&6.2\\
 & Top-5 & 22.8&20.5&22.3&20.7&18.3&22.2&19.7&22.0&17.6&28.3&21.4\\
ATM & Top-1 & 10.5&7.1&11.9&14.7&7.0&11.1&16.1&15.0&4.9&20.5&11.9\\
 & Top-5 & 26.8&24.8&33.8&39.4&23.9&35.8&43.5&40.3&22.7&46.5&33.8\\
UBP & Top-1 & 11.5&15.5&9.8&13.0&8.8&11.7&10.2&12.2&15.5&16.0&12.4\\
 & Top-5 & 29.7&40.0&27.0&32.3&33.8&31.0&23.8&32.2&40.5&43.5&33.4\\
NeuroBridge & Top-1 & 23.2&21.2&13.2&17.0&14.5&25.0&15.3&20.1&13.7&27.2&19.0\\
 & Top-5 & 52.4&49.3&36.5&45.3&37.7&55.0&45.1&44.9&36.5&56.3&45.9\\
Shallow Alignment & Top-1 & 24.6&31.3&11.4&19.9&19.0&24.1&18.6&17.6&23.3&34.6&22.4\\
 & Top-5 & 54.7&61.5&31.1&48.8&45.5&49.8&51.6&46.7&54.9&63.2&50.8\\
SAMGA & Top-1 & \textbf{36.3}&\textbf{42.2}&\textbf{24.7}&\textbf{34.0}&\textbf{30.0}&\textbf{35.2}&\textbf{35.1}&\textbf{28.7}&\textbf{28.1}&\textbf{49.6}&\textbf{34.4}\\
 & Top-5 & \textbf{69.8}&\textbf{71.7}&\textbf{50.7}&\textbf{66.7}&\textbf{64.3}&\textbf{67.7}&\textbf{61.0}&\textbf{59.5}&\textbf{57.0}&\textbf{79.4}&\textbf{64.8}\\
\bottomrule
\end{tabular}}
\end{table*}

\subsection{Experimental Setup}

All experiments are implemented in PyTorch and conducted on an NVIDIA RTX 3090 GPU. SAMGA is trained for 60 epochs using AdamW with a batch size of 1024, an initial learning rate of $1\times10^{-4}$, and a weight decay of $1\times10^{-4}$. SAMGA results are averaged over five random seeds.

For THINGS-EEG, we use 17 occipital and parietal electrodes in the intra-subject setting and all 63 EEG channels in the inter-subject setting. Image representations are extracted from the frozen InternViT-6B-448px-V2.5 vision encoder \citep{r53}. Five intermediate layers, $\{20,24,28,32,36\}$, are used for multi-granularity target construction. Following the intermediate-depth preference identified by Shallow Alignment \citep{r22}, the global routing logits are initialized around layer 28. The principal implementation settings are summarized in Table~1.

\begin{table}[!htbp]
\centering
\caption*{Table 1. Main implementation settings.}
\label{tab:implementation_settings_visual}
\begin{tabular}{@{}>{\raggedright\arraybackslash}p{0.33\columnwidth}>{\raggedright\arraybackslash}p{0.59\columnwidth}@{}}
\toprule
Item & Setting\\
\midrule
Selected visual layers & $\{20,24,28,32,36\}$\\
Routing initialization & Initial logits $[-2,-1,0,-1,-2]$, centered at layer 28\\
Training epochs & 60\\
Optimizer & AdamW\\
Batch size & 1024\\
Initial learning rate & $1\times10^{-4}$\\
Weight decay & $1\times10^{-4}$\\
\bottomrule
\end{tabular}
\end{table}

For comparison, we include representative EEG-based visual decoding methods covering different alignment strategies. NICE \citep{r13} establishes contrastive EEG--image retrieval in a pretrained visual space, while ATM \citep{r16} combines EEG--visual alignment with a guided generative prior. UBP \citep{r17} adapts image-side information through uncertainty-aware visual degradation, and NeuroBridge \citep{r18} incorporates cognitive priors and bidirectional semantic alignment. Shallow Alignment \citep{r22} is the closest method to our granularity formulation and replaces conventional final-layer supervision with an intermediate visual representation.

\subsection{Comparison with Representative Methods}

\subsubsection{Results on THINGS-EEG}

Table~\ref{tab:main_results} compares SAMGA with representative EEG-to-image retrieval methods under both intra-subject and inter-subject settings.

Under intra-subject evaluation, SAMGA achieves an average Top-1 accuracy of 91.3\% and Top-5 accuracy of 98.8\%. Compared with Shallow Alignment, the strongest competing method, SAMGA improves Top-1 accuracy from 82.6\% to 91.3\%, corresponding to an 8.7 percentage-point gain. The Top-1 improvement is observed across all ten subjects, while the average Top-5 accuracy further increases from 97.7\% to 98.8\%.

The performance advantage is more pronounced under leave-one-subject-out evaluation. SAMGA achieves 34.4\% Top-1 and 64.8\% Top-5 accuracy, compared with 22.4\% and 50.8\% for Shallow Alignment, yielding improvements of 12.0 and 14.0 percentage points, respectively. SAMGA also obtains the highest Top-1 and Top-5 accuracy for every held-out subject in Table~\ref{tab:main_results}. These gains under subject shift are consistent with subject-dependent calibration around a shared routing prior.

\subsubsection{Results on THINGS-MEG}

Table~\ref{tab:meg_results} reports the retrieval results on THINGS-MEG. Under intra-subject evaluation, SAMGA achieves 48.1\% Top-1 and 72.5\% Top-5 accuracy, outperforming Shallow Alignment by 4.3 and 3.3 percentage points, respectively.

Under leave-one-subject-out evaluation, SAMGA reaches 6.1\% Top-1 and 16.6\% Top-5 accuracy, compared with 3.8\% and 13.3\% for Shallow Alignment. The gains under both evaluation protocols indicate that the proposed target-construction strategy remains effective for MEG-based visual retrieval.

\begin{table}[!htbp]
\centering
\caption{Comparison with representative methods on THINGS-MEG.}
\label{tab:meg_results}
\footnotesize
\renewcommand{\arraystretch}{1.02}
\setlength{\tabcolsep}{1.5pt}
\begin{tabular}{@{}llrrrrr@{}}
\toprule
\multicolumn{7}{c}{\textbf{Intra-subject: train and test on one subject}}\\
\midrule
Method & Metric & Sub1 & Sub2 & Sub3 & Sub4 & Avg.\\
\midrule
UBP & Top-1 & 15.0 & 46.0 & 27.3 & 18.5 & 26.7\\
 & Top-5 & 38.0 & 80.5 & 59.0 & 43.5 & 55.2\\
NeuroBridge & Top-1 & 16.5 & 53.7 & 40.4 & 18.1 & 32.2\\
 & Top-5 & 41.6 & 85.3 & 73.2 & 43.1 & 60.8\\
Shallow Alignment & Top-1 & 23.6 & 70.5 & 54.2 & 26.8 & 43.8\\
 & Top-5 & 46.3 & 90.4 & 85.0 & 54.9 & 69.2\\
SAMGA & Top-1 & \textbf{26.3} & \textbf{77.5} & \textbf{61.3} & \textbf{27.1} & \textbf{48.1}\\
 & Top-5 & \textbf{53.6} & \textbf{94.9} & \textbf{86.2} & \textbf{55.1} & \textbf{72.5}\\
\midrule
\multicolumn{7}{c}{\textbf{Inter-subject: leave one subject out for test}}\\
\midrule
Method & Metric & Sub1 & Sub2 & Sub3 & Sub4 & Avg.\\
\midrule
UBP & Top-1 & 2.0 & 1.5 & 2.7 & 2.5 & 2.2\\
 & Top-5 & 5.7 & 17.2 & 10.5 & 8.0 & 10.4\\
NeuroBridge & Top-1 & 4.3 & 3.6 & 3.0 & 2.5 & 3.4\\
 & Top-5 & 13.1 & 15.6 & 11.2 & \textbf{11.3} & 12.8\\
Shallow Alignment & Top-1 & 1.4 & 6.7 & 5.6 & 1.7 & 3.8\\
 & Top-5 & 8.7 & 18.2 & 18.5 & 7.6 & 13.3\\
SAMGA & Top-1 & \textbf{5.6} & \textbf{9.0} & \textbf{6.3} & \textbf{3.7} & \textbf{6.1}\\
 & Top-5 & \textbf{13.5} & \textbf{23.7} & \textbf{21.5} & 7.6 & \textbf{16.6}\\
\bottomrule
\end{tabular}
\end{table}

\subsection{Ablation Studies}

\subsubsection{Multi-Granularity Target Construction}

To assess the contribution of multi-granularity target construction, we compare three visual supervision schemes: a fixed single-layer target at layer 28, uniform fusion of the five selected visual layers, and the learned fusion used by SAMGA. This comparison separates the effect of incorporating hierarchical information from that of adaptively learning its composition.

As shown in Table~\ref{tab:fusion_ablation}, the fixed layer-28 target obtains 87.8\% Top-1 and 98.4\% Top-5 accuracy. Uniformly combining the five visual layers improves Top-1 accuracy to 89.1\%, indicating that representations from different visual depths provide complementary information. Learning the fusion weights further increases Top-1 accuracy to 91.3\%. The progression from fixed single-layer supervision to uniform fusion and learned fusion shows that the gain arises not only from incorporating multiple representation depths, but also from adaptively constructing the visual target from them.

\FloatBarrier

\begin{table}[!htbp]
\centering
\caption{Ablation of multi-granularity visual-target construction under the intra-subject setting.}
\label{tab:fusion_ablation}
\begin{tabular}{lcc}
\toprule
Target construction & Top-1 & Top-5\\
\midrule
Fixed single layer (Layer 28) & 87.8 & 98.4\\
Uniform fusion & 89.1 & 98.7\\
Learned fusion & \textbf{91.3} & \textbf{98.8}\\
\bottomrule
\end{tabular}
\end{table}

\subsubsection{Coarse-to-Fine Alignment}

Table~\ref{tab:twostage_ablation} examines the optimization design of the proposed coarse-to-fine alignment. Replacing the staged procedure with one-stage training decreases Top-1 accuracy from 91.3\% to 87.5\%, showing that separating the two optimization stages contributes substantially to retrieval performance.

Stage-specific optimization further improves the staged procedure. Removing the reduced learning-rate strategy in the fine stage decreases Top-1 accuracy to 89.0\%, whereas keeping the shared encoder trainable during fine alignment results in 91.0\%. These results indicate that the main improvement originates from the coarse-to-fine optimization itself, with the learning-rate transition providing a further gain and fixing the shared encoder contributing a smaller refinement.

\begin{table}[!htbp]
\centering
\caption{Ablation of the coarse-to-fine alignment strategy under the intra-subject setting.}
\label{tab:twostage_ablation}
\begin{tabular}{lcc}
\toprule
Variant & Top-1 & Top-5\\
\midrule
One-stage training & 87.5 & 98.0\\
W/o stage-specific LR & 89.0 & 98.3\\
W/o shared-encoder freezing & 91.0 & 98.7\\
SAMGA & \textbf{91.3} & \textbf{98.8}\\
\bottomrule
\end{tabular}
\end{table}

\subsubsection{Projector Design}

\begin{figure}[!htbp]
\centering
\includegraphics[width=0.92\linewidth]{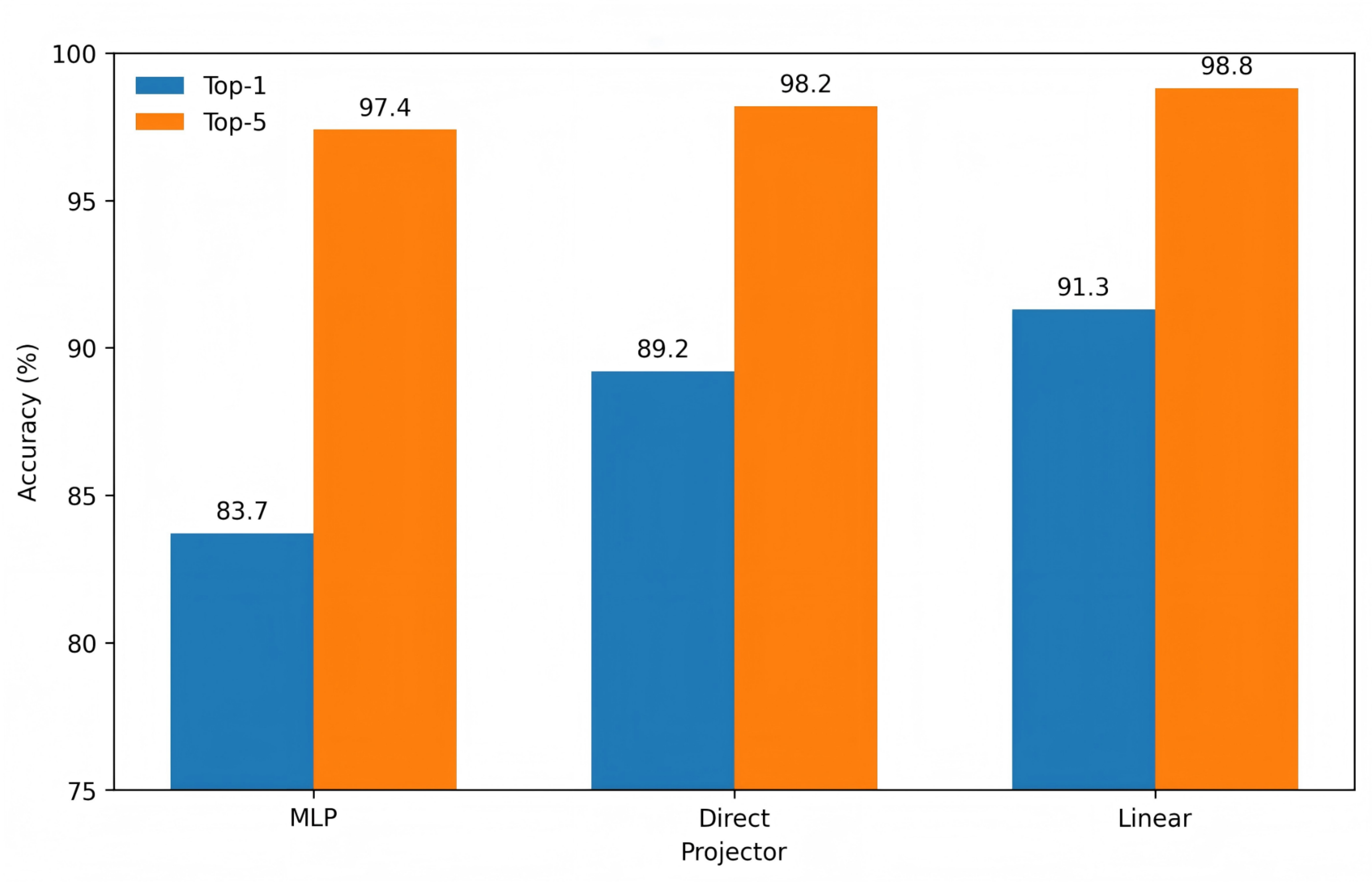}
\caption{Comparison of different projector designs for EEG--visual alignment.}
\label{fig:projector}
\end{figure}

We further examine the projector used to map hierarchical visual representations into the alignment space. As shown in Fig.~\ref{fig:projector}, the linear projector achieves the highest retrieval accuracy, outperforming both direct alignment without an additional projector and the MLP-based alternative. This result indicates that a lightweight learned transformation is sufficient for mapping hierarchical visual representations into a common alignment space, whereas additional nonlinear mapping does not provide further benefit in this setting.

\subsection{Representation and Granularity Analysis}

\subsubsection{Semantic Structure of the Shared Representation}

\begin{figure}[!htbp]
\centering
\includegraphics[width=0.96\linewidth]{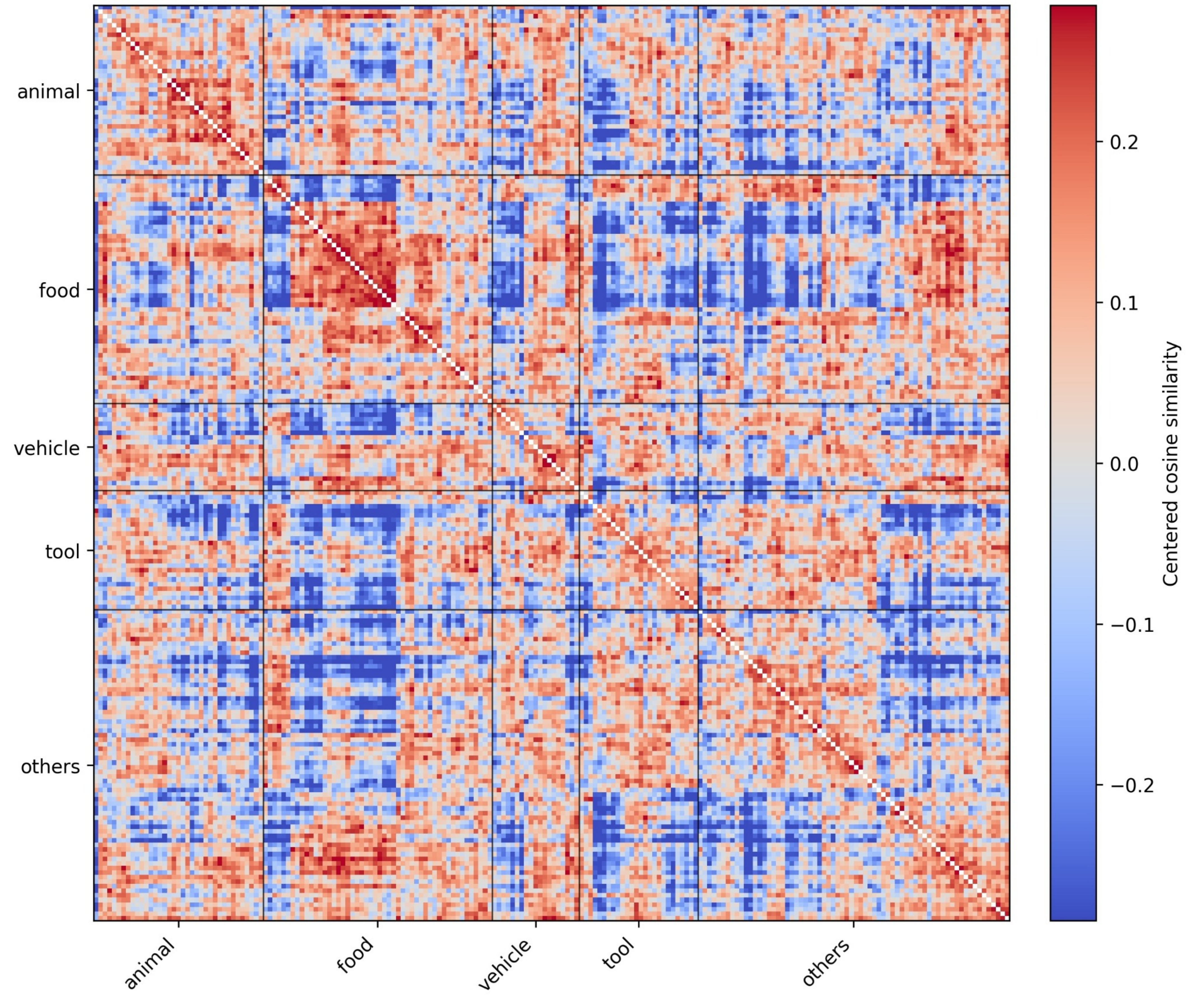}
\caption{Inter-subject concept similarity matrix of EEG concept embeddings on the 200 test concepts. Concepts are reordered by coarse semantic categories, and similarity values are centered to highlight relative semantic structure.}
\label{fig:concept_similarity}
\end{figure}

We next examine the semantic organization of EEG representations learned under the inter-subject setting. Following NICE \citep{r13}, the 200 test concepts are grouped into five coarse semantic categories---animal, food, vehicle, tool, and others. Fig.~\ref{fig:concept_similarity} shows the pairwise concept-similarity matrix after reordering the concepts by these categories. Clear local structure can be observed within several semantic groups, particularly for animal and food concepts, whereas tool and the heterogeneous others group exhibit more dispersed similarity patterns.

\begin{figure}[!htbp]
\centering
\includegraphics[width=0.84\linewidth]{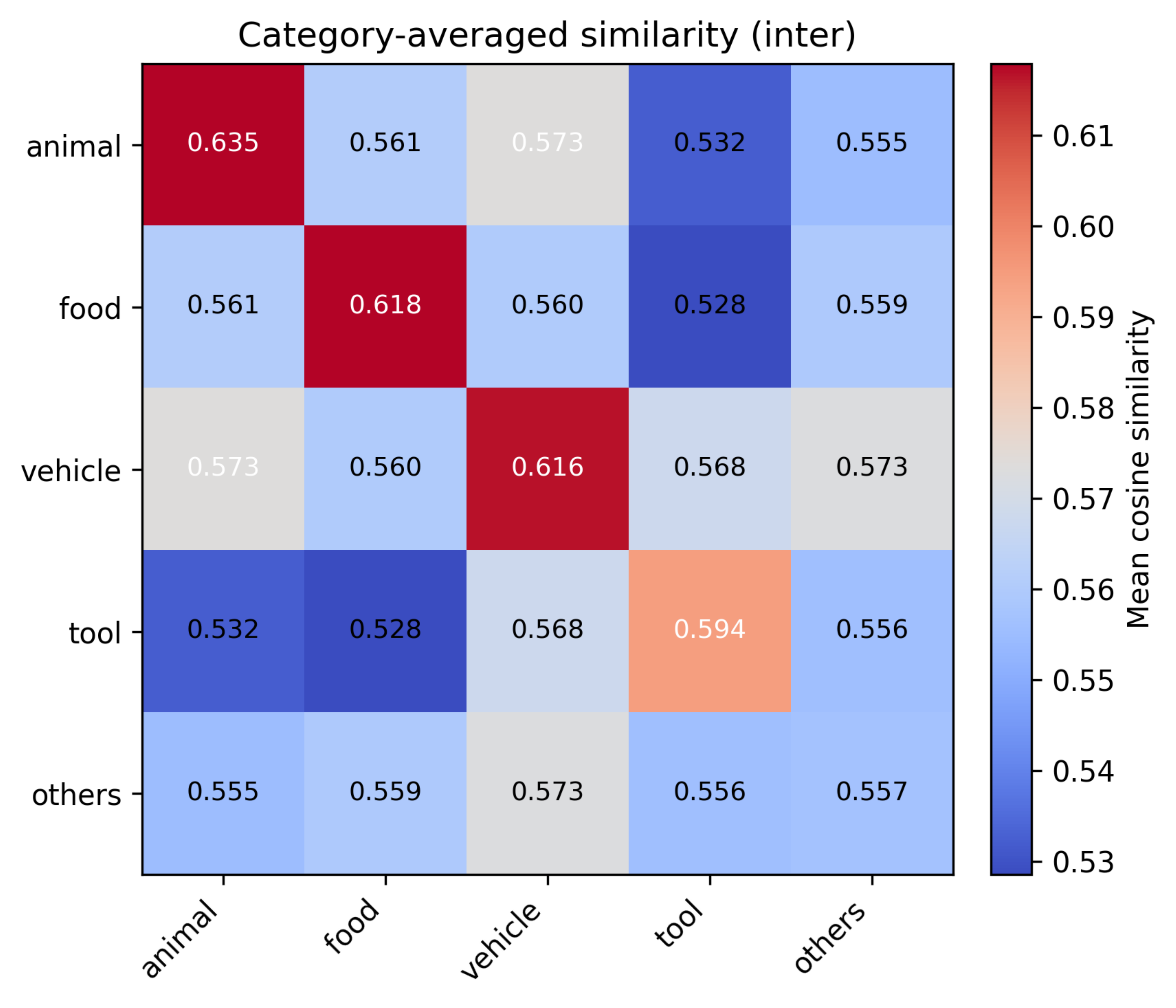}
\caption{Category-level mean similarity matrix under the inter-subject setting. Diagonal entries denote within-category similarity, whereas off-diagonal entries denote between-category similarity.}
\label{fig:category_similarity}
\end{figure}

Fig.~\ref{fig:category_similarity} summarizes the same representation structure at the category level. Animal, food, vehicle, and tool each exhibit their highest mean similarity on the corresponding diagonal entry, while the heterogeneous others group shows weaker within-category compactness. These patterns indicate that the shared EEG representation retains category-related semantic organization across held-out subjects.

\subsubsection{Subject-Dependent Visual-Depth Preference}

\begin{figure}[!htbp]
\centering
\includegraphics[width=0.94\linewidth]{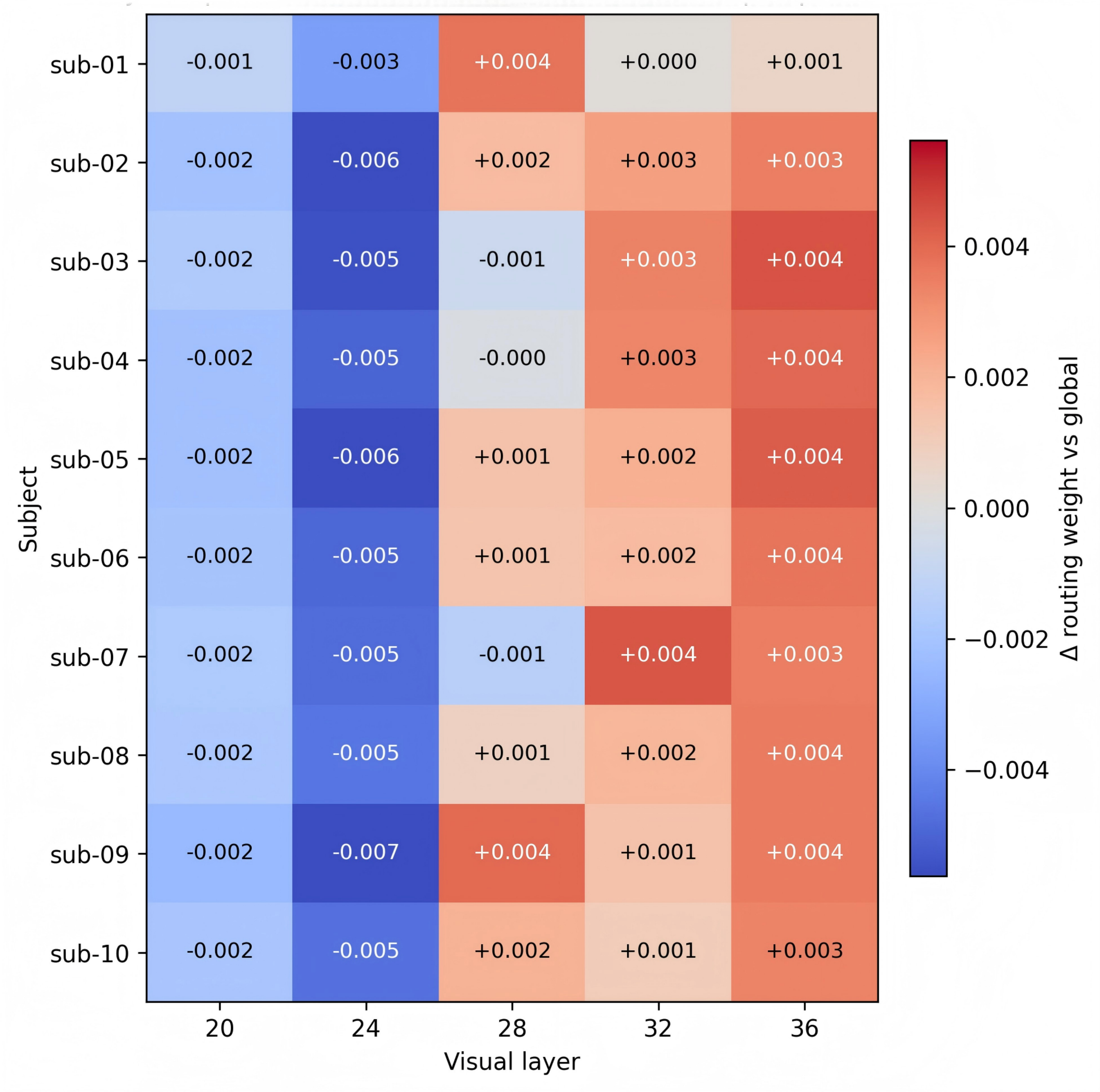}
\caption{Subject-specific deviation from the global visual-depth preference. Positive values indicate higher routing weight than the global reference, and negative values indicate lower routing weight.}
\label{fig:subject_deviation}
\end{figure}

Fig.~\ref{fig:subject_deviation} visualizes the learned subject-specific deviations from the global routing distribution. Each entry represents the difference between a subject-conditioned routing weight and the corresponding global routing weight at the same visual depth.

The residual corrections are small in magnitude and exhibit structured variation across both visual depth and subject, consistent with their role as lightweight calibration around the shared global routing reference. Layer 24 receives consistently negative corrections across subjects, whereas layers 32 and 36 generally receive positive corrections. Layer 28 exhibits greater subject-dependent variation, with positive deviations for some subjects and negative deviations for others. The learned routing therefore retains a common granularity reference while allowing subject information to redistribute a limited portion of the visual-depth weighting during training.

\subsubsection{Category-Wise Visual-Depth Analysis}

We further examine whether the preferred visual depth varies with semantic content. To isolate the contribution of visual depth, a separate fixed-target model is trained for each candidate visual layer while keeping the remaining training and evaluation settings unchanged.

\begin{figure}[!htbp]
\centering
\includegraphics[width=0.96\linewidth]{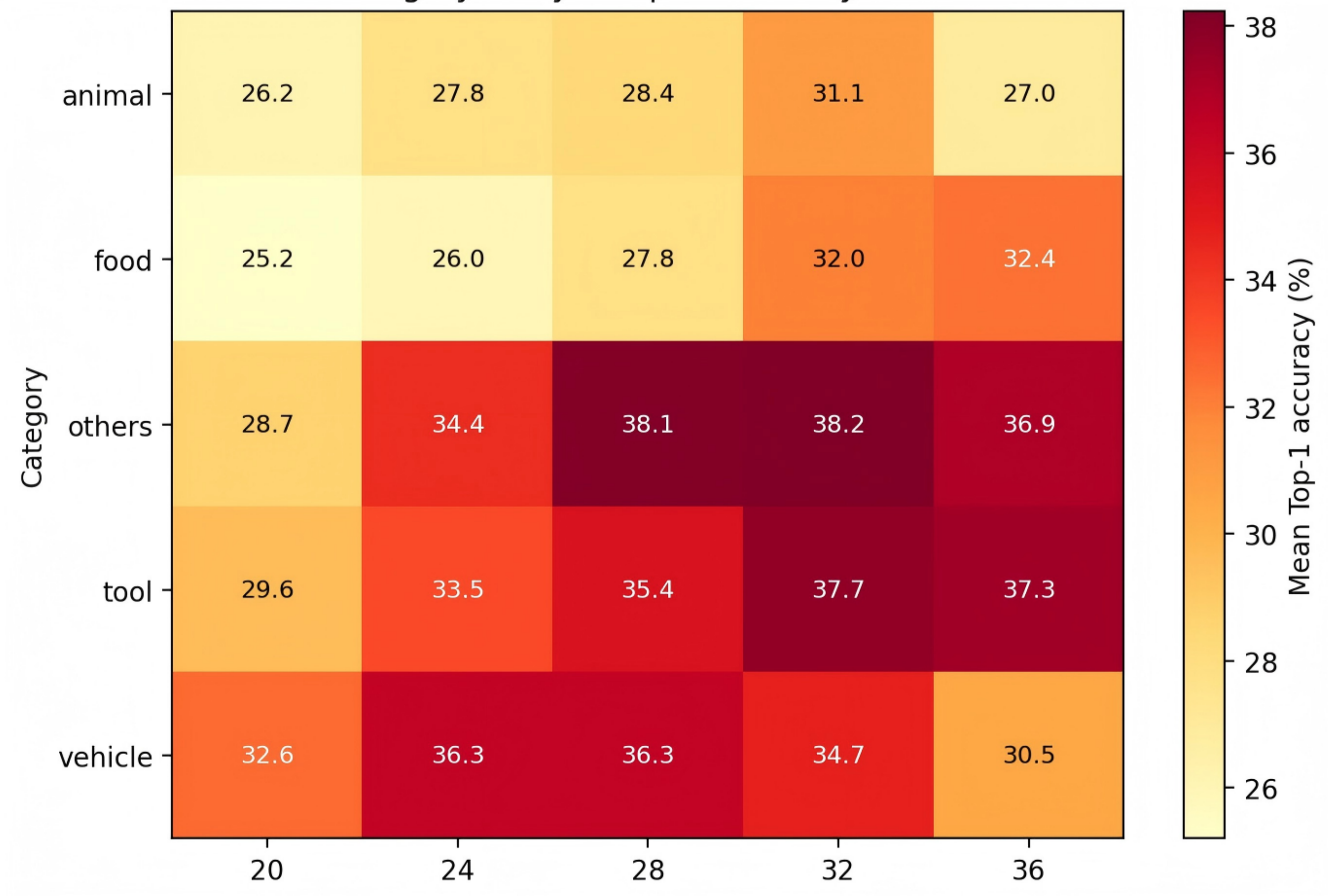}
\caption{Category-wise Top-1 retrieval accuracy across visual layers under the inter-subject setting.}
\label{fig:coarse_category}
\end{figure}

At the coarse semantic level, Fig.~\ref{fig:coarse_category} shows that the highest Top-1 retrieval accuracy is obtained at different representation depths across categories. Animal, tool, and others achieve their best performance at layer 32, food favors layer 36, whereas vehicle performs best at the shallower intermediate layers 24 and 28. A single fixed representation depth therefore does not provide the strongest supervision uniformly across semantic groups.

\begin{figure}[!htbp]
\centering
\includegraphics[width=0.96\linewidth]{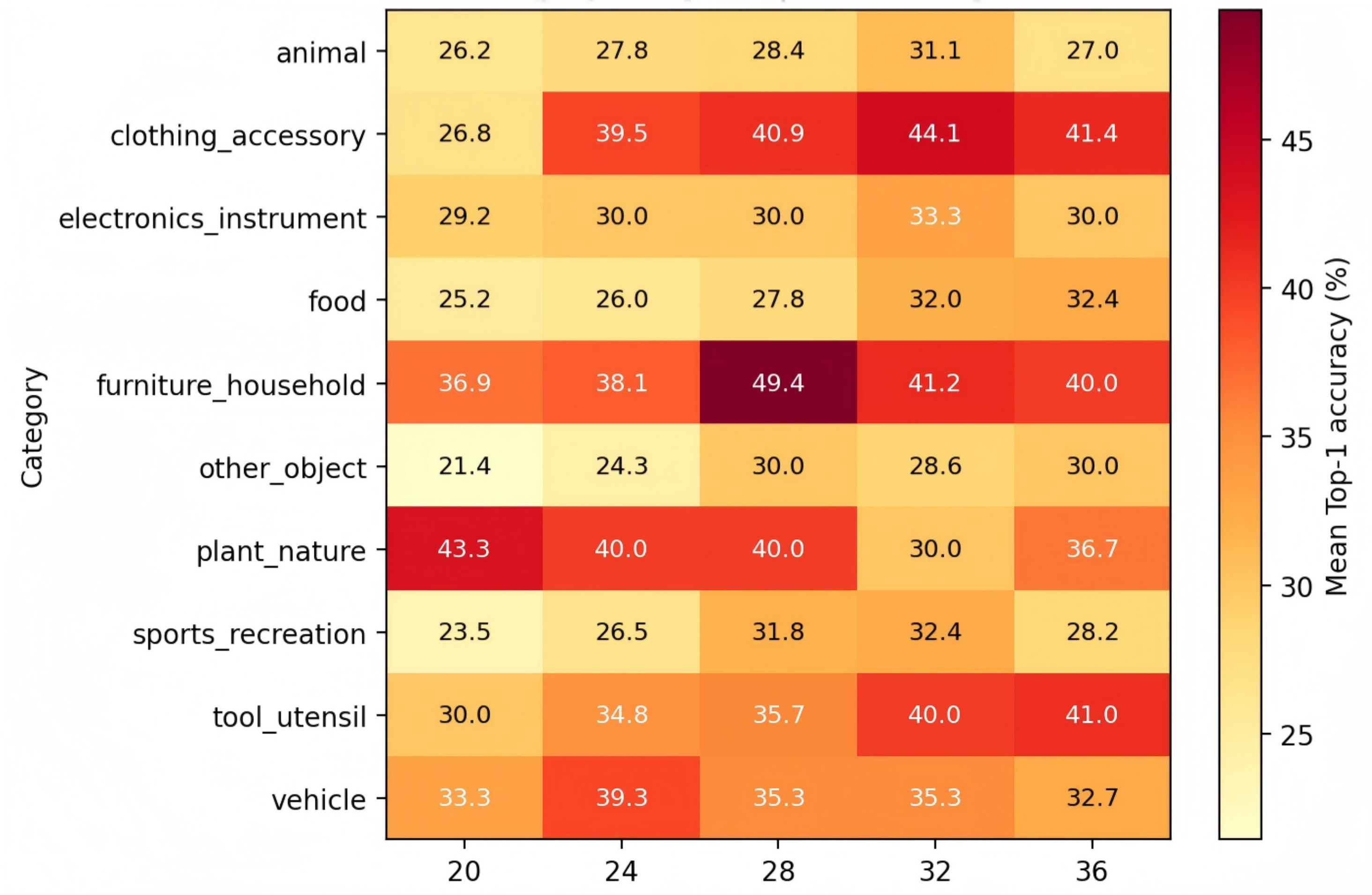}
\caption{Fine-grained category-wise Top-1 retrieval accuracy across visual layers under the inter-subject setting.}
\label{fig:fine_category}
\end{figure}

For a finer-grained analysis, we further subdivide the 200 test concepts into ten semantic groups based on their object labels. Fig.~\ref{fig:fine_category} shows that the resulting preference spans the full candidate hierarchy: plant-related concepts favor layer 20, vehicle favors layer 24, furniture and household concepts peak at layer 28, several categories achieve their strongest performance at layer 32, while food and tool-related concepts favor layer 36. The category-dependent pattern provides complementary evidence that EEG-compatible visual supervision is distributed across multiple levels of the visual hierarchy rather than concentrated at a universally optimal depth.

\FloatBarrier

\section{Discussion}

A central finding of this study is that visual supervision should not be treated as a fixed endpoint of EEG--visual alignment. Existing approaches have largely focused on improving neural representation learning or the objective used to associate EEG with a pretrained visual space. Our results instead show that how the visual target is constructed can substantially affect retrieval performance. This observation extends the granularity perspective introduced by Shallow Alignment \citep{r22}: while selecting an appropriate intermediate representation is already preferable to relying on highly abstract final-layer features, the present results indicate that a single fixed intermediate depth does not fully capture the visual information that is useful for EEG decoding.

The multi-granularity analyses provide complementary evidence for this point. Uniformly combining several intermediate visual representations improves over the fixed layer-28 target, while learned fusion yields a further improvement, indicating that the benefit arises from both hierarchical information and its adaptive composition. The category-wise analysis provides a related view of the same phenomenon. Different semantic groups achieve their highest retrieval accuracy at different visual depths, with the preferred layer spanning the candidate hierarchy rather than concentrating at one universally optimal representation. These observations suggest that the utility of visual representations for neural supervision varies across levels of abstraction. Importantly, this result concerns the compatibility between EEG representations and visual-network features; it should not be interpreted as a direct correspondence between individual semantic categories and specific stages of biological visual processing.

Subject variability introduces a second dimension to this target construction problem. The larger performance gains observed under leave-one-subject-out evaluation indicate that the advantage of SAMGA becomes particularly evident when the neural distribution shifts across individuals. At the same time, the learned subject residuals remain small in magnitude and exhibit structured variations across subjects and visual depths. This behavior is consistent with the intended global--residual formulation: the global component captures the dominant granularity tendency, whereas subject information provides lightweight calibration during training rather than defining an independent target for each individual. Consequently, subject-aware optimization does not require subject-dependent inference. More broadly, this suggests that subject information in multi-subject neural decoding need not be restricted to subject-specific neural mappings; it can also calibrate shared supervision during training without requiring subject identity at deployment.

The coarse-to-fine results further indicate that target construction and cross-modal optimization play complementary roles. Replacing the staged procedure with one-stage optimization produces the largest degradation among the alignment variants, whereas the stage-specific learning-rate transition and fixing of the shared encoder provide additional refinements. This pattern supports separating shared-space organization from instance-level retrieval refinement. During coarse alignment, distribution-level alignment and pairwise correspondence jointly establish a common EEG--visual geometry; fine alignment then emphasizes retrieval discrimination while preserving the shared transformation established in the preceding stage. Taken together, these results suggest that the advantage of the coarse-to-fine strategy stems not only from the alignment objectives themselves, but also from how their roles are organized over the course of optimization.

Taken together, these findings broaden the role of visual supervision in neural-to-visual retrieval. Rather than treating pretrained visual representations as a fixed destination for neural mapping, SAMGA treats target construction as part of the alignment problem itself. This target-side perspective also provides an alternative use of subject information, namely calibrating supervision rather than only modifying neural representations or learning subject-specific mappings. The results on THINGS-MEG further support this perspective, showing that the same design remains applicable when neural responses are measured with a different non-invasive recording modality. These results do not imply direct transfer between EEG and MEG, since the models are trained separately for the two datasets, but they indicate that the benefit of granularity-aware target construction is not restricted to a single neural measurement setting.

Several limitations define the current scope of SAMGA. First, granularity variation is modeled at the subject level, implicitly assuming that the preferred visual supervision for an individual is relatively stable across trials. Trial-dependent cognitive state, attention, and response variability may lead to finer-grained variation in EEG--visual compatibility that is not represented by the current routing formulation. Extending the router toward sample- or state-adaptive calibration could allow such variability to be modeled at a finer resolution. Second, SAMGA learns how to combine a predefined set of visual depths rather than discovering the candidate hierarchy itself. Jointly identifying informative depths and their contributions may provide a more flexible form of target construction. Third, the present evaluation focuses on zero-shot object retrieval within the THINGS stimulus space. Extending the framework to more complex natural scenes, continuous visual perception, and additional neural datasets would help determine how broadly the observed granularity patterns extend beyond object-level retrieval.

\section{Conclusion}

This work revisits zero-shot EEG-to-image retrieval from the perspective of visual supervision construction. We proposed Subject-Aware Multi-Granularity Alignment (SAMGA), which constructs EEG-compatible visual targets from hierarchical visual representations through a global granularity prior and subject-dependent residual calibration, and organizes the resulting EEG--visual correspondence using coarse-to-fine alignment. On THINGS-EEG, SAMGA improves Top-1 retrieval accuracy over the strongest competing method by 8.7 percentage points under intra-subject evaluation and 12.0 percentage points under leave-one-subject-out evaluation. Ablation studies further show that learned multi-granularity target construction outperforms both fixed-layer and uniform fusion strategies, while representation analyses reveal structured variations in visual-depth preference across subjects and semantic groups. Results on THINGS-MEG further support the applicability of the proposed target-side formulation beyond EEG. Overall, these findings indicate that robust neural--visual alignment depends not only on how neural representations are encoded and optimized, but also on how the visual supervision target itself is constructed. This perspective complements existing encoder and objective design and offers an alternative direction for neural--visual retrieval across subjects.

\FloatBarrier


\begin{thebibliography}{99}
\setlength{\bibsep}{0.55\baselineskip}

\bibitem[Ahmed et al.(2021)]{r42}
Ahmed, H., Wilbur, R. B., Bharadwaj, H. M., \& Siskind, J. M. (2021). Object classification from randomized EEG trials. In \textit{Proceedings of the IEEE/CVF Conference on Computer Vision and Pattern Recognition} (pp. 3845--3854).

\bibitem[Benchetrit et al.(2024)]{r49}
Benchetrit, Y., Banville, H., \& King, J.-R. (2024). Brain decoding: Toward real-time reconstruction of visual perception. In \textit{International Conference on Learning Representations}.

\bibitem[Bralina et al.(2026)]{r50}
Bralina, S., Yazici, A., Guan, C., \& Lee, M.-H. (2026). Adaptive bottleneck transformer for multimodal EEG, audio, and vision fusion. \textit{Expert Systems with Applications, 312}, 131487. \url{https://doi.org/10.1016/j.eswa.2026.131487}

\bibitem[Chen et al.(2026)]{r52}
Chen, J., Pi, D., Gao, F., Ma, C., \& Chen, Y. (2026). MI-DGHCL: Motor imagery EEG domain generalization via hyperbolic contrastive learning. \textit{Expert Systems with Applications, 312}, 131477. \url{https://doi.org/10.1016/j.eswa.2026.131477}

\bibitem[Chen et al.(2024)]{r53}
Chen, Z., Wang, W., Cao, Y., Liu, Y., Gao, Z., Cui, E., Zhu, J., Ye, S., Tian, H., Liu, Z., Gu, L., Wang, X., Li, Q., Ren, Y., Chen, Z., Luo, J., Wang, J., Jiang, T., Wang, B., \ldots{} Wang, W. (2024). Expanding performance boundaries of open-source multimodal models with model, data, and test-time scaling [Preprint]. \textit{arXiv}. \url{https://arxiv.org/abs/2412.05271}

\bibitem[Dai et al.(2025)]{r35}
Dai, Y., Yao, Z., Song, C., Zheng, Q., Mai, W., Peng, K., Lu, S., Ouyang, W., Yang, J., \& Wu, J. (2025). MindAligner: Explicit brain functional alignment for cross-subject visual decoding from limited fMRI data. In \textit{Proceedings of the 42nd International Conference on Machine Learning} (Vol. 267, pp. 12214--12228). PMLR.

\bibitem[Du et al.(2023)]{r6}
Du, C., Fu, K., Li, J., \& He, H. (2023). Decoding visual neural representations by multimodal learning of brain-visual-linguistic features. \textit{IEEE Transactions on Pattern Analysis and Machine Intelligence, 45}(9), 10760--10777. \url{https://doi.org/10.1109/TPAMI.2023.3263181}

\bibitem[Du et al.(2026)]{r22}
Du, Y., Dai, S., Song, Y., Thompson, P. M., Tang, H., \& Zhan, L. (2026). Deep models, shallow alignment: Uncovering the granularity mismatch in neural decoding [Preprint]. \textit{arXiv}. \url{https://arxiv.org/abs/2601.21948}

\bibitem[Gifford et al.(2022)]{r10}
Gifford, A. T., Dwivedi, K., Roig, G., \& Cichy, R. M. (2022). A large and rich EEG dataset for modeling human visual object recognition. \textit{NeuroImage, 264}, 119754. \url{https://doi.org/10.1016/j.neuroimage.2022.119754}

\bibitem[Gong et al.(2025)]{r34}
Gong, Z., Zhang, Q., Bao, G., Zhu, L., Xu, R., Liu, K., Hu, L., \& Miao, D. (2025). MindTuner: Cross-subject visual decoding with visual fingerprint and semantic correction. \textit{Proceedings of the AAAI Conference on Artificial Intelligence, 39}(13), 14247--14255. \url{https://doi.org/10.1609/aaai.v39i13.33560}

\bibitem[Gretton et al.(2012)]{r38}
Gretton, A., Borgwardt, K. M., Rasch, M. J., Sch\"olkopf, B., \& Smola, A. J. (2012). A kernel two-sample test. \textit{Journal of Machine Learning Research, 13}, 723--773.

\bibitem[Grootswagers et al.(2022)]{r40}
Grootswagers, T., Zhou, I., Robinson, A. K., Hebart, M. N., \& Carlson, T. A. (2022). Human EEG recordings for 1,854 concepts presented in rapid serial visual presentation streams. \textit{Scientific Data, 9}, 3. \url{https://doi.org/10.1038/s41597-021-01102-7}

\bibitem[G\"u\c{c}l\"u \& van Gerven(2015)]{r21}
G\"u\c{c}l\"u, U., \& van Gerven, M. A. J. (2015). Deep neural networks reveal a gradient in the complexity of neural representations across the ventral stream. \textit{Journal of Neuroscience, 35}(27), 10005--10014. \url{https://doi.org/10.1523/JNEUROSCI.5023-14.2015}

\bibitem[Haxby et al.(2011)]{r23}
Haxby, J. V., Guntupalli, J. S., Connolly, A. C., Halchenko, Y. O., Conroy, B. R., Gobbini, M. I., Hanke, M., \& Ramadge, P. J. (2011). A common, high-dimensional model of the representational space in human ventral temporal cortex. \textit{Neuron, 72}(2), 404--416. \url{https://doi.org/10.1016/j.neuron.2011.08.026}

\bibitem[Hebart et al.(2023)]{r36}
Hebart, M. N., Contier, O., Teichmann, L., Rockter, A. H., Zheng, C. Y., Kidder, A., Corriveau, A., Vaziri-Pashkam, M., \& Baker, C. I. (2023). THINGS-data, a multimodal collection of large-scale datasets for investigating object representations in human brain and behavior. \textit{eLife, 12}, e82580. \url{https://doi.org/10.7554/eLife.82580}

\bibitem[Horikawa \& Kamitani(2017)]{r20}
Horikawa, T., \& Kamitani, Y. (2017). Generic decoding of seen and imagined objects using hierarchical visual features. \textit{Nature Communications, 8}, 15037. \url{https://doi.org/10.1038/ncomms15037}

\bibitem[Kamitani \& Tong(2005)]{r4}
Kamitani, Y., \& Tong, F. (2005). Decoding the visual and subjective contents of the human brain. \textit{Nature Neuroscience, 8}(5), 679--685. \url{https://doi.org/10.1038/nn1444}

\bibitem[Kwon et al.(2025)]{r51}
Kwon, B.-H., Lee, M., \& Lee, S.-W. (2025). Functional connectivity guided deep neural network for decoding high-level visual imagery. \textit{Expert Systems with Applications, 294}, 128734. \url{https://doi.org/10.1016/j.eswa.2025.128734}

\bibitem[Li et al.(2024)]{r16}
Li, D., Wei, C., Li, S., Zou, J., \& Liu, Q. (2024). Visual decoding and reconstruction via EEG embeddings with guided diffusion. In \textit{Advances in Neural Information Processing Systems} (Vol. 37, pp. 102822--102864). \url{https://doi.org/10.52202/079017-3266}

\bibitem[Li et al.(2025)]{r45}
Li, Y., Kang, Z., Gong, S., Dong, W., Zeng, W., Yan, H., Siok, W. T., \& Wang, N. (2025). Neural-MCRL: Neural multimodal contrastive representation learning for EEG-based visual decoding. In \textit{2025 IEEE International Conference on Multimedia and Expo (ICME)} (pp. 1--6). \url{https://doi.org/10.1109/ICME59968.2025.11210130}

\bibitem[Long et al.(2015)]{r39}
Long, M., Cao, Y., Wang, J., \& Jordan, M. I. (2015). Learning transferable features with deep adaptation networks. In \textit{Proceedings of the 32nd International Conference on Machine Learning} (Vol. 37, pp. 97--105). PMLR.

\bibitem[McCartney et al.(2022)]{r43}
McCartney, B., Devereux, B., \& Mart\'inez-del-Rinc\'on, J. (2022). A zero-shot deep metric learning approach to Brain--Computer Interfaces for image retrieval. \textit{Knowledge-Based Systems, 246}, 108556. \url{https://doi.org/10.1016/j.knosys.2022.108556}

\bibitem[Qu et al.(2026)]{r47}
Qu, T., Yang, Z., \& Zhang, Q. (2026). NeuroVision: EEG-to-image reconstruction via progressive neural encoding and cross-modal distillation. \textit{Expert Systems with Applications, 312}, 131526. \url{https://doi.org/10.1016/j.eswa.2026.131526}

\bibitem[Radford et al.(2021)]{r12}
Radford, A., Kim, J. W., Hallacy, C., Ramesh, A., Goh, G., Agarwal, S., Sastry, G., Askell, A., Mishkin, P., Clark, J., Krueger, G., \& Sutskever, I. (2021). Learning transferable visual models from natural language supervision. In \textit{Proceedings of the 38th International Conference on Machine Learning} (Vol. 139, pp. 8748--8763). PMLR.

\bibitem[Robinson et al.(2023)]{r5}
Robinson, A. K., Quek, G. L., \& Carlson, T. A. (2023). Visual representations: Insights from neural decoding. \textit{Annual Review of Vision Science, 9}, 313--335. \url{https://doi.org/10.1146/annurev-vision-100120-025301}

\bibitem[Scotti et al.(2024)]{r33}
Scotti, P. S., Tripathy, M., Torrico, C., Kneeland, R., Chen, T., Narang, A., Santhirasegaran, C., Xu, J., Naselaris, T., Norman, K. A., \& Abraham, T. M. (2024). MindEye2: Shared-subject models enable fMRI-to-image with 1 hour of data. In \textit{Proceedings of the 41st International Conference on Machine Learning} (Vol. 235, pp. 44038--44059). PMLR.

\bibitem[Shen et al.(2019)]{r25}
Shen, G., Horikawa, T., Majima, K., \& Kamitani, Y. (2019). Deep image reconstruction from human brain activity. \textit{PLoS Computational Biology, 15}(1), e1006633. \url{https://doi.org/10.1371/journal.pcbi.1006633}

\bibitem[Song et al.(2024)]{r13}
Song, Y., Liu, B., Li, X., Shi, N., Wang, Y., \& Gao, X. (2024). Decoding natural images from EEG for object recognition. In \textit{International Conference on Learning Representations}.

\bibitem[Song et al.(2025)]{r14}
Song, Y., Wang, Y., He, H., \& Gao, X. (2025). Recognizing natural images from EEG with language-guided contrastive learning. \textit{IEEE Transactions on Neural Networks and Learning Systems, 36}(9), 15896--15910. \url{https://doi.org/10.1109/TNNLS.2025.3562743}

\bibitem[Spampinato et al.(2017)]{r41}
Spampinato, C., Palazzo, S., Kavasidis, I., Giordano, D., Souly, N., \& Shah, M. (2017). Deep learning human mind for automated visual classification. In \textit{Proceedings of the IEEE Conference on Computer Vision and Pattern Recognition} (pp. 6809--6817).

\bibitem[Sun et al.(2026)]{r48}
Sun, K., Miao, X., Zhai, B., Duan, H., \& Long, Y. (2026). Decoding visual neural representations by multimodal with dynamic balancing. \textit{Expert Systems with Applications, 297}, 129382. \url{https://doi.org/10.1016/j.eswa.2025.129382}

\bibitem[Wang et al.(2025)]{r19}
Wang, J., Zhang, L., Lin, H., Liu, Q., Huang, G., Li, Z., Liang, Z., \& Wu, X. (2025). NeuroCLIP: Brain-inspired prompt tuning for EEG-to-image multimodal contrastive learning [Preprint]. \textit{arXiv}. \url{https://arxiv.org/abs/2511.09250}

\bibitem[Wang et al.(2024)]{r32}
Wang, S., Liu, S., Tan, Z., \& Wang, X. (2024). MindBridge: A cross-subject brain decoding framework. In \textit{Proceedings of the IEEE/CVF Conference on Computer Vision and Pattern Recognition} (pp. 11333--11342).

\bibitem[Wu et al.(2025)]{r17}
Wu, H., Li, Q., Zhang, C., He, Z., \& Ying, X. (2025). Bridging the vision-brain gap with an uncertainty-aware blur prior. In \textit{Proceedings of the IEEE/CVF Conference on Computer Vision and Pattern Recognition} (pp. 2246--2257). \url{https://doi.org/10.1109/CVPR52734.2025.00215}

\bibitem[Xiong et al.(2025)]{r15}
Xiong, D., Hu, L., Jin, J., Ding, Y., Tan, C., Zhang, J., \& Tian, Y. (2025). Interpretable cross-modal alignment network for EEG visual decoding with algorithm unrolling. \textit{IEEE Transactions on Neural Networks and Learning Systems, 36}(11), 19894--19908. \url{https://doi.org/10.1109/TNNLS.2025.3592646}

\bibitem[Xu et al.(2025)]{r46}
Xu, T., Yu, L., Zheng, Y., \& Huang, S. (2025). BrainVision: Cross-domain EEG decoding for visual content retrieval and reconstruction. \textit{Neuroscience, 584}, 190--205. \url{https://doi.org/10.1016/j.neuroscience.2025.07.047}

\bibitem[Yamins et al.(2014)]{r37}
Yamins, D. L. K., Hong, H., Cadieu, C. F., Solomon, E. A., Seibert, D., \& DiCarlo, J. J. (2014). Performance-optimized hierarchical models predict neural responses in higher visual cortex. \textit{Proceedings of the National Academy of Sciences of the United States of America, 111}(23), 8619--8624. \url{https://doi.org/10.1073/pnas.1403112111}

\bibitem[K. Zhang et al.(2025)]{r44}
Zhang, K., He, L., Jiang, X., Lu, W., Wang, D., \& Gao, X. (2025). CognitionCapturer: Decoding visual stimuli from human EEG signal with multimodal information. \textit{Proceedings of the AAAI Conference on Artificial Intelligence, 39}(13), 14486--14493. \url{https://doi.org/10.1609/aaai.v39i13.33587}

\bibitem[T. Zhang et al.(2025)]{r30}
Zhang, T., Wan, F., Sun, K., Miao, X., Sun, Y., Shao, M., \& Long, Y. (2025). Dynamic convolution and graph-coupled attention for cross-subject EEG--vision decoding. In \textit{Proceedings of the 36th British Machine Vision Conference} (Paper 509). BMVA.

\bibitem[Zhang et al.(2026)]{r18}
Zhang, W., Wang, S., Su, Y., Li, X., Zhang, C., \& Zhong, S. (2026). NeuroBridge: Bio-inspired self-supervised EEG-to-image decoding via cognitive priors and bidirectional semantic alignment. \textit{Proceedings of the AAAI Conference on Artificial Intelligence, 40}(21), 18028--18036. \url{https://doi.org/10.1609/aaai.v40i21.38863}

\bibitem[Zhu et al.(2025)]{r9}
Zhu, S., Ye, Z., Ai, Q., \& Liu, Y. (2025). CrossPT-EEG: A benchmark for cross-participant and cross-time generalization of EEG-based visual decoding [Preprint, Version 2]. \textit{arXiv}. \url{https://arxiv.org/abs/2406.07151v2}

\end{thebibliography}
\end{document}